\documentclass{article}

\usepackage{arxiv}

\usepackage[utf8]{inputenc} % allow utf-8 input
\usepackage[T1]{fontenc}    % use 8-bit T1 fonts
\usepackage{hyperref}       % hyperlinks
\usepackage{url}            % simple URL typesetting
\usepackage{booktabs}       % professional-quality tables
\usepackage{amsfonts}       % blackboard math symbols
\usepackage{nicefrac}       % compact symbols for 1/2, etc.
\usepackage{microtype}      % microtypography
\usepackage{lipsum}		% Can be removed after putting your text content
\usepackage{graphicx}
\usepackage{doi}

\usepackage[toc,page]{appendix}
\usepackage{float}
\usepackage{booktabs}
\usepackage[dvipsnames]{xcolor}

\title{Hybrid Long Document Summarization Using C2F-FAR and ChatGPT: A Practical Study}

\author{ 
{Guang Lu, Sylvia B. Larcher} \\
	Institute of Communication and Marketing\\
	Lucerne University of Applied Sciences and Arts\\
	Zentralstrasse 9, 6002 Lucerne, Switzerland \\
	\texttt{guang.lu@hslu.ch, sylvia.bendel@hslu.ch} \\
	\And
    {Tu Tran} \\
	getAbstract AG\\
	Alpenquai 12, 6005 Lucerne, Switzerland\\
	\texttt{tu.tran@getAbstract.com} \\
}

\renewcommand{\headeright}{Hybrid Long Document Summarization Using C2F-FAR and ChatGPT}
\renewcommand{\undertitle}{}
\renewcommand{\shorttitle}{}

\begin{document}
\maketitle

\begin{abstract}
Text summarization is an important downstream natural language processing (NLP) task that challenges both the understanding and generation capabilities of language models. Thanks to large language models (LLMs) and techniques for fine-tuning models in machine learning, significant progress has been made in automatically summarizing short texts such as news articles, often leading to very satisfactory machine-generated results. In contrast, summarizing long documents still remains a major challenge. This is partly due to the complex nature of contextual information in long texts, but also due to the lack of open-source benchmarking datasets and the corresponding evaluation frameworks that can be used to develop and test model performance. In this work, we use ChatGPT, the latest breakthrough in the field of NLP and LLMs, together with the extractive summarization model C2F-FAR (Coarse-to-Fine Facet-Aware Ranking) to propose a hybrid extraction and summarization pipeline for long documents such as business articles and books. We work with the world-renowned company getAbstract AG and leverage their expertise and experience in professional book summarization. A practical study has shown that machine-generated summaries can perform at least as well as human-written summaries when evaluated using current automated evaluation metrics. However, a closer examination of the texts generated by ChatGPT through human evaluations has shown that there are still critical issues in terms of text coherence, faithfulness, and style. Overall, our results show that the use of ChatGPT is a very promising but not yet mature approach for summarizing long documents and can at best serve as an inspiration for human editors. We anticipate that our work will inform NLP researchers about the extent to which ChatGPT's capabilities for summarizing long documents overlap with practitioners' needs. Further work is needed to test the proposed hybrid summarization pipeline, in particular involving GPT-4, and to propose a new evaluation framework tailored to the task of summarizing long documents.
\end{abstract}

% keywords can be removed

\section{Introduction}
\label{sec:1}
Text summarization is a long-standing task in the field of natural language processing (NLP), and the corresponding techniques can be distinguished as extractive and abstractive \cite{ElKassas2021Automatic}. Extractive summarization identifies key sentences that can convey the most important information in the original text and uses these as summaries, while abstractive summarization captures the semantic meanings of a document by first understanding the text and then constructing new sentences as summaries. Both approaches have their merits and are sometimes even combined into a hybrid extraction and summarization pipeline to achieve better model performance \cite{Pilault2020On,Cajueiro2023A}. 

Automatic summarization of short texts, such as news articles (about 650 words per document), has made significant progress thanks to the rapid development of large language models (LLMs), the large benchmarking datasets, and a variety of model fine-tuning techniques in machine learning \cite{Zhao2023A,Yang2023Harnessing,Zhang2023Benchmarking}. By fine-tuning pre-trained LLMs or directly tuning a specially developed language model to the given datasets, the model can usually achieve very good summarization performance comparable to human-written summaries \cite{Fabbri2021SummEval}. However, summarizing long documents remains a daunting challenge \cite{Koh2022An}. This is simply because long documents, such as business articles and books, can contain much longer text (and thus lower information noise ratio), more complex contextual information, and more complicated hierarchical structures. In addition, most currently developed LLMs cannot accept long text inputs that exceed their token limit. Although some special techniques have been developed, such as top-down and bottom-up inference \cite{Pang2022Long} and recursive summarization with human feedback \cite{Stiennon2020Learning,Wu2021Recursively,Ouyang2022Training}, to enable the model to read long documents and even books in particular, the performance of the model still depends heavily on fine-tuning or human intervention. Finally, there is also a lack of systematic and expert assessment of the quality of machine-generated summaries in general \cite{Kryściński2019Neural}. 

Since the release of OpenAI's ChatGPT in late November 2022, numerous studies have been conducted using ChatGPT's performance for various language tasks, including text summarization \cite{Zhang2023A,Cao2023A,Bang2023A,Zhang2023One}. Sometimes even seemingly contradictory results were reported, which is consistent with the complex nature of the task. For example, Soni and Wade \cite{Soni2023Comparing} studied the performance of ChatGPT on abstractive summarization tasks. Their results suggest that humans cannot easily distinguish machine-produced text from human-produced text. Yang et al. \cite{Yang2023Exploring} used ChatGPT for query-based and aspect-based text summarization tasks and found that ChatGPT's performance on ROUGE \cite{Lin2004ROUGE} scores was comparable to conventionally fine-tuned models. Zhang et al. \cite{Zhang2023Extractive} thoroughly investigated the performance of ChatGPT on extractive summarization using various benchmarking datasets. They showed that ChatGPT is still inferior to state-of-the-art models subjected to supervised fine-tuning. However, they pointed out that an extraction and generation pipeline using ChatGPT is a better solution than the baseline abstractive summarization models in terms of faithfulness. Wang et al. \cite{Wang2023Cross} even tested the newly released GPT-4 using ChatGPT on cross-lingual summarization tasks and reported that although ChatGPT tends to produce long summaries showing many details of the text, it achieves excellent zero-shot performance compared to the leading cross-lingual summarization models. Finally, there is also research on applying ChatGPT to specific domains, such as radiology report summarization \cite{Ma2023ImpressionGPT}, or using ChatGPT as additional evaluators for text quality assessment \cite{Luo2023Chatgpt,Liu2023GPTEval,Gao2023Human,Wang2023Is}. Although it seems promising to incorporate ChatGPT in automatic text summarization, it remains unclear how the model performs in summarizing long documents such as business articles and books, and what impact ChatGPT has on practical summarization services in industry.

In this work, we attempt to fill this research gap by using ChatGPT for summarizing long documents. However, due to the existing limitation on the number of tokens that ChatGPT allows as user input (4,096 tokens for GPT-3.5 and 8,000 tokens for the standard GPT-4), an additional extractive summarization approach, i.e., Coarse-to-Fine Facet-Aware Ranking (C2F-FAR) \cite{Liang2022An}, is applied to first extract important sentences from the given texts. We show how the proposed hybrid extraction and summarization pipeline can be used for summarizing long documents, even books. We demonstrate the good performance of ChatGPT in summary generation, given the main automated evaluation metrics. However, when manually reviewing the generated summaries and comparing them with human-written summaries, we find that ChatGPT can still cause obvious problems in terms of text coherence, faithfulness, and style. The results of this practical study provide researchers and practitioners with insight into the real capabilities of ChatGPT in summarizing long documents, and highlight the main challenges that the text summarization technique needs to solve in the future. Therefore, the contributions of this paper can be summarized as follows:
\begin{itemize}
\item {We presented the first study using a hybrid text summarization pipeline with C2F-FAR and ChatGPT for summarizing long documents.}
\item {We studied the performance of ChatGPT in summarizing long business articles and books highlighted by automated evaluation metrics and human annotations.}
\item {We provided insights into the strengths and weaknesses of the proposed summarization pipeline and pointed out the next key challenges that LLMs like ChatGPT need to solve in summarizing long documents.}
\end{itemize}

\section{Related Work}
\label{sec:2}
This section provides a brief overview of long document summarization models, benchmarking datasets, and summary evaluation approaches. It is not intended to be a complete survey of all related studies. Rather, it provides an overview of recent representative developments and shows why harnessing the zero-shot power of LLMs like ChatGPT could be a viable way to summarize long documents, especially for practical purposes.

\subsection{Techniques for Summarizing Long Documents}
\label{sec:2.1}
Many extractive and abstractive summarization models have been developed that aim to summarize long documents of up to 10,000 words \cite{Koh2022An}. While most of these models are powerful, they are trained based on given datasets and need to be fine-tuned or completely re-trained for new text domains, making the application of these models tedious in practice. Therefore, the zero-shot  capability of LLMs is seen as a promising alternative for summarizing long documents \cite{Bubeck2023Sparks,Liu2023Summary}.  

\subsubsection{Extractive Summarization}
\label{sec:2.1.1}
Existing approaches for extractive summarization are mainly based on Transformer \cite{Zhang2022HEGEL,Liu2021HETFORMER,Ruan2022HiStruct,Cho2022Toward}, graph \cite{Zhang2022HEGEL,Bian2022GoSum,Xie2022GRETEL}, and reinforcement learning \cite{Bian2022GoSum,Gu2021MemSum} models, as well as their combinations \cite{Zhang2022HEGEL,Bian2022GoSum}. Although these models show promising results in summarizing relatively long documents (e.g., 3,000 to 8,000 words), they need to be trained and fine-tuned based on the given new datasets, which is time-consuming and hinders the direct application of these models in practice. For example, Zhang et al. \cite{Zhang2022HEGEL} proposed HEGEL, which stands for HypErGraph transformer for Extractive Long document summarization, and used the so-called hyperedges to model high-order cross-sentence relations in different perspectives, such as section structure, latent topic, and keyword coreference. Since the proposed hyperedge can connect an arbitrary number of vertices, the model is able to effectively learn sentence embeddings and capture their interdependence using the designed hypergraph transformation layers. The test of HEGEL on two benchmarking datasets (arXiv and PubMed \cite{Cohan2018A}) showed almost always the best results in terms of ROUGE \cite{Lin2004ROUGE} scores compared to other recent models. A more recent study by Bian et al. \cite{Bian2022GoSum} proposed GoSum, which aims to train the agent's action based on its state in a reinforcement learning environment to evaluate and select sentences and produce an extractive summary. Specifically, the states of the sentences were encoded using graph neural networks (GNNs), which capture the hierarchical structure of the document via a heterogeneous graph that treats sentences and sections as nodes. This enables GoSum to intelligently select sentences and create summaries that take into account local, global, and historical sentence information. The performance of GoSum on the arXiv and PubMed datasets \cite{Cohan2018A}, based on the measurement of ROUGE \cite{Lin2004ROUGE} scores, showed the best results compared to the other state-of-the-art extractive and abstractive summarization models.

\subsubsection{Abstractive Summarization}
\label{sec:2.1.2}
The modern abstractive summarization approaches for long documents (e.g., up to 16,000 words) are almost exclusively based on Transformer and its derivatives \cite{Cao2022HIBRIDS,Zhang2021SummN,Mao2021DYLE,Phang2022Investigating,He2022zCode}. Again, despite the complex model architecture that ensures good summarization  assessed by the existing automated scoring metrics, in practice the models need to be re-trained with the new datasets to fully exploit their summarization capabilities. In other words, the features of the documents and summaries that led to the original trained model may affect its performance in the new dataset. An example would be the GOVREPORT dataset \cite{Huang2021Efficient}, where sometimes content from the beginning of the documents appears more frequently in the human-written summary, which may not be desirable for the other summarization tasks. An interesting model Summ$^{N}$ \cite{Zhang2021SummN} has been proposed based on a multi-stage split-then-summarize framework, where the summarization process is divided into several coarse stages and a fine stage. For each coarse stage, the model needs to segment and combine the parts of the document with the summary and let the backbone model generate temporary summaries for each segment. These generated summaries are combined to repeat the process so that the long document is condensed step by step until the text length in the final stage matches the token limit of the backbone model. It has been claimed that Summ$^{N}$ can be applied to summarize texts of arbitrary length by adjusting the number of stages. However, determining the number of stages depends solely on the number of tokens that can be fed into the backbone model without considering semantic segmentation within the original documents. Besides these works, few studies have investigated the approach to summarize even longer documents such as books (e.g., more than 100,000 words). Pang et al. \cite{Pang2022Long} proposed the so-called top-down and bottom-up inference framework, which allows learning robust token representations to capture both detailed local textual information and long-range dependency between different chapters of the document. It was found that after training, this model can achieve state-of-the-art summarization results based on ROUGE \cite{Lin2004ROUGE} scores even for book datasets \cite{Kryscinski2021BookSum}, but requires less than 1\% of the parameters required for other GPT-3-based models. In contrast to this approach, Wu et al. \cite{Wu2021Recursively} used reinforcement learning to recursively summarize long documents (e.g., books) considering human feedback in various intermediate steps. Their method decomposed the entire text into multiple sections of reasonable length. They trained machine learning models for these sections as well as human-written summaries for these sections. Based on the generated summaries, human demonstrations were collected to develop a reward mechanism to optimize the model's performance in summarizing. These steps were repeated until the entire document was summarized. The model has been reported to provide high quality summaries, in some cases even comparable to human-written summaries. In addition, humans in the loop can help reduce the problem of hallucinations that is common in automatic text generation \cite{Ji2023Survey}. However, involving humans in the process requires professional knowledge when providing feedback to the model and could be a complex process in practice. 

\subsection{Datasets and Evaluation Methods for Long Documents Summarization}
\label{sec:2.2}
To support the training of machine learning models for summarizing long documents, suitable datasets are needed, but they are relatively scarce. However, there are some recent efforts to fill the gap in this area. The two previously mentioned datasets, arXiv and PubMed \cite{Cohan2018A}, were collected from the two scientific repositories, arXiv.org and PubMed.com, and each contains approximately 100,000 to 200,000 documents with an average length of a document ranging from 3,000 to 5,000 words. Ladhak et al. \cite{Ladhak2020Exploring} focused on novel chapter summarization tasks and proposed a novel chapter dataset with 8,088 collected chapter and summary pairs from 79 books. The average length of the chapters is about 5,000 words and the average number of words in the summaries is 370. Huang et al. \cite{Huang2021Efficient} presented a GOVREPORT dataset containing 19,466 U.S. government reports with an average length of 9,409 words and the corresponding expert-written summary with an average length of 553 words. This dataset has become a standard benchmark for the development of long document summarization models \cite{Pang2022Long,Cao2022HIBRIDS,Zhang2021SummN,Koh2022How}. Kryściński et al. \cite{Kryscinski2021BookSum} published a large dataset BookSum that includes a variety of long document types such as novels, plays, and stories, as well as the corresponding human-written summaries of 142,753 paragraphs, 12,293 chapters, and 436 books. They also provided strong base summarization models for other researchers working on this dataset. Recently, Wang et al. \cite{Wang2022SQuALITY} collected a dataset in which the documents are 625 public-domain short stories, approximately 5,200 words in length, and the summaries specifically focus on answering certain questions. This provides useful additional perspectives for summarizing long documents. 

So far, ROUGE \cite{Lin2004ROUGE} has remained the simplest metric for evaluating the relevance of machine-generated summaries compared to human-written summaries. However, a high ROUGE score does not necessarily indicate factual summaries. There are more other evaluation metrics that can be distinguished as either reference-based methods such as BERTScore \cite{Zhang2019BERTScore} and MoverScore \cite{Zhao2019MoverScore} or reference-free methods such as BLANC \cite{Vasilyev2020Fill} and ESTIME \cite{Vasilyev2021ESTIME,Vasilyev2021Consistency}, which focus more on the semantic overlaps between the machine-generated and human-written summaries, how the generated summary can support the textual understanding of the original document, and the factual correspondence between the texts. To support human evaluation of generated text, Vig et al. \cite{Vig2021SummVis} implemented a visualization tool SummVis that allows humans to conveniently compare a source document, a reference summary, and a generated summary by highlighting words and phrases that match between them. Human evaluation helps identify factual inconsistencies and text coherence issues that are not easily found based on automated metrics alone \cite{Gehrmann2022Repairing}. Readers interested in other approaches to evaluating text summarization results can find them at \cite{Fabbri2021SummEval,Koh2022An}.

\section{Data and Methods}
\label{sec:3}
In this practical study, we tested a hybrid text extraction and summarization pipeline to summarize long documents. For this purpose, we used business articles and books and for manual quality comparison a corresponding subset of getAbstract summaries written by humans. Since we assumed that the tested pipeline should easily support human editors in creating summaries, we did not train the existing models mentioned in Section \ref{sec:2.1} on our dataset. Instead, we leveraged the performance of the unsupervised extractive summarization model C2F-FAR (Coarse-to-Fine Facet-Aware Ranking) \cite{Liang2022An} and ChatGPT to identify the strengths and weaknesses of this summarization pipeline and the key challenges for its use in productions. The use of ChatGPT is subject to OpenAI's strict privacy policy\footnotemark{}\footnotetext{https://openai.com/policies/privacy-policy} and all associated data is used for research purposes only. 

\subsection{Data}
\label{sec:3.1}
Based in Switzerland, getAbstract AG is a world-renowned company whose primary mission is to provide high-quality, expertly written summaries for a wide range of English-language business books, articles, videos and business reports so that readers can quickly acquire new knowledge in the business world. In this work, getAbstract provided summaries in plain text. We randomly selected and cleaned a set of articles and books to test the proposed summarization pipeline. A subset of the names of these articles and books can be found in Appendix \ref{app:A}.

We used two methods to prepare the cleaned text for the summarization. The first method is based on fully automated tools. Here, the PDF texts were extracted using the Python package Parsr\footnotemark{}, which can effectively exclude figures, tables and other visual elements in the original document. The tool generates the plain text file and additionally a MARKDOWN file (text with syntax for formatting heading, subheading, bold word, etc.), a JSON file (all detailed information of the extraction) and a CSV file (all table data in the PDF is saved as a CSV file)\footnotetext{https://github.com/axa-group/Parsr}. We also wrote a Python program to remove other unnecessary elements in the text, such as website and email addresses and non-ACSII characters. This method was applied to the selected business articles, since they are not as long as books, resulting in a reasonable computation time. The second method is based on manual and automatic cleaning with Python applied to the selected business books. Each copied PDF text in Notepad++ was first manually cleaned with Regex to remove all obvious unwanted patterns within the document. For example, the beginning pages up to the first chapter and the ending pages after the last chapter of the book were removed. Then we replaced all matches to the given Regex with a blank. These given Regex include, for example, table and figure captions and website addresses. Next, we used the developed Python program to search the text and find other patterns that we need to remove from the text (e.g., page header, footnote, chapter title, page number, author name). More precisely, we have developed a logic based on the format of a well-written book to concatenate meaningful lines in the text starting from the last line. We assumed that the first and the last line should be valid and meaningful lines of the text. We then went backwards and checked whether the previous line makes sense up to the current line, i.e., whether they belong to a continuous text. Basically, we checked the five lines before the current line and determined: (1) If there is a previous end of sentence in reverse order within the five lines (i.e., period + space + capital letter or period + end of line), we append that previous line to the current line. (2) If no sentence ending is found in the previous five lines, we look for the first line in backward order whose first character is an uppercase letter, and consider this line as the sentence beginning of the current line. (3) If neither of the above is true, the previous lines are discarded and considered as noise coming from tables, figures, headings, headers, footers, and page numbers. After cleaning and concatenating all lines, we also removed non-ASCII characters, several contiguous spaces, and very short sentences within the text with a length of less than four words.  

Using the above methods, we obtained a relatively cleaned dataset of 20 articles and 20 books. The 20 articles have an average word count of 5,603 and the 20 books have an average word count of 67,031. Although the number of documents is small, it is sufficient to test the summarization pipeline that uses the unsupervised learning of C2F-FAR and the zero-shot capability of ChatGPT.  

\subsection{Methods}
\label{sec:3.2}

\subsubsection{C2F-FAR}
\label{sec:3.2.1}
\begin{figure}[H]
    \centering
    \includegraphics[width=6.5in]{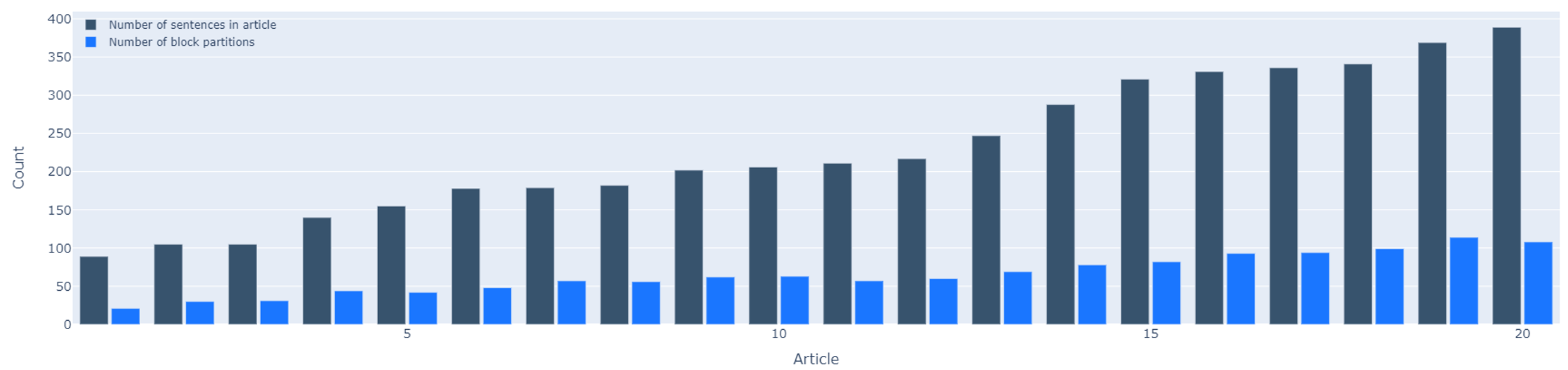}
    \caption{Testing the C2F-FAR default setting on the 20 articles in our dataset. The number of block partitions generally increases with the number of sentences in the article, which shows that the model can automatically capture the semantic blocks in texts.}
    \label{Fig1}
\end{figure}

C2F-FAR is an unsupervised method that can automatically select sentences from a document by considering the semantic similarity between different blocks of text and between different sentences. The advantage is that this method can be applied directly without having to be trained on a large amount of data. However, there are several parameters that can be tuned for optimal performance on the targeted dataset. The idea behind the C2F-FAR method is straightforward. All individual sentences within the document must first be encoded into vectors using some pre-trained language models as their semantic representations. Then, an analysis process is performed in two stages. In the coarse stage, the document is divided into semantic blocks based on a measure of similarity between adjacent sentences. It is assumed that sentences in the same semantic block should focus on the same topic and a semantically different sentence from the previous ones starts a new semantic block. A developed centrality estimator is used to filter out unimportant blocks and keep only the important ones according to the given ratio. In the fine stage, a relevance estimator is used to calculate the similarity of each sentence within the block to the block representation, which is determined by averaging all sentence representation vectors in the same block. Only the highest scoring sentences are retained. The remaining sentences from all blocks are run through another centrality estimator to select those that are eligible for extractive summarization. The parameters we set in the C2F-FAR method can affect, for example, the number of semantic blocks, the threshold above which the sentences in a block must be selected, and the output length of the summary. Figure \ref{Fig1} takes the cleaned 20 articles to apply the default C2F-FAR setting and shows how the number of block partitions generally increases with the number of sentences in an article.

\subsubsection{ChatGPT}
\label{sec:3.2.2}
\begin{figure}[H]
    \centering
    \includegraphics[width=6.5in]{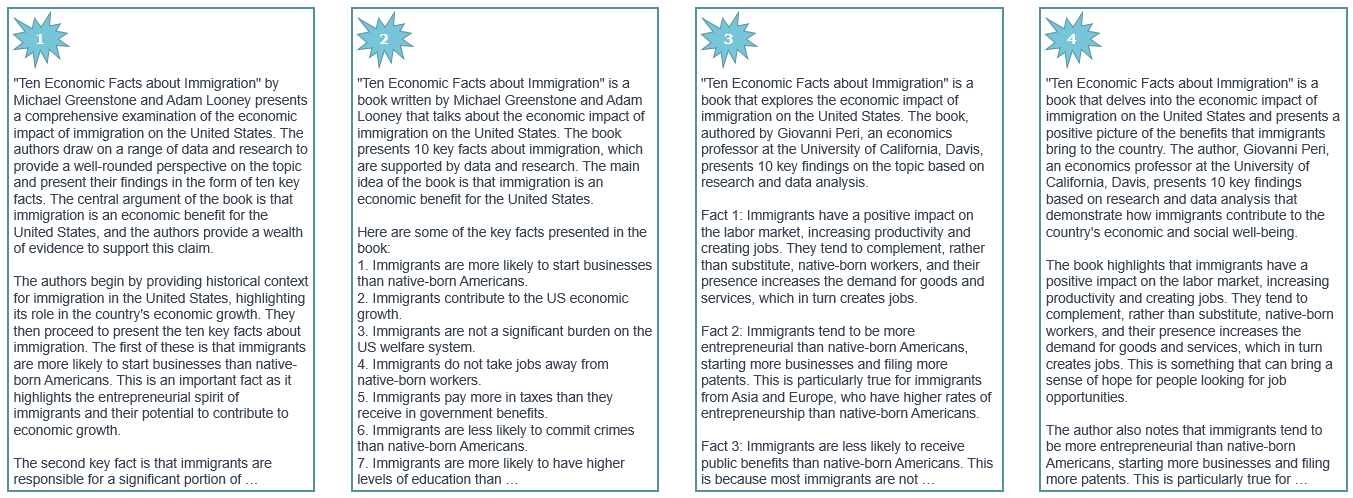}
    \caption{Comparison of ChatGPT summaries of the long article ``Ten Economic Facts about Immigration'' on the basis of four required styles: (1) professional, (2) easy to understand, (3) fact-oriented, and (4) emotion-oriented. The prompt influences the output text to some extent.}
    \label{Fig2}
\end{figure}

ChatGPT is a powerful tool for understanding and generating languages. In this work, we used ChatGPT for summarization based on the extractive summaries from C2F-FAR to study the problems in summarizing long documents. We designed the prompts as a simple two-step process. For example, for the given business articles, we first asked ChatGPT to ``Please summarize the following text in your own words in about 25 sentences.'' If the generated summaries are too short, we further asked ChatGPT to generate 25 sentences or add more details to the summary. We found that this two-step prompting resulted in satisfactory summary length in most cases when compared to human-written summaries. However, if we repeatedly ask ChatGPT to provide more details, it may provide text that is unrelated to the original document. For the given business books, we repeated the above procedure, but had to enter the sentences extracted from C2F-FAR into ChatGPT in chunks so that ChatGPT could summarize step by step. The chunk was simply defined as about three pages in a Word document. We were aware that this may cause coherence problems in the summaries produced by ChatGPT. However, we believe that this could be an effective way to get around the token limitation of the ChatGPT input and assume that the model output can still give us insight into the overall quality of the texts. All of these summarization tests were performed between January and April 2023 using the standard version of ChatGPT based on GPT-3.5, with the exception of a final whole book summarization test which was additionally conducted using GPT-4. Figure \ref{Fig2} shows an example where we asked ChatGPT to summarize the article ``Ten Economic Facts about Immigration'' by M. Greenstone and A. Looney in different styles. In this simple example, we directly asked ChatGPT to create a summary based on the given article name instead of copying the entire article text into the user interface. The prompt reads, ``Please summarize the article (name of article) in approximately 500 words using a (description of style) style.'' The influence of the prompts on the output texts can be clearly seen. 

\section{Results and Discussion}
\label{sec:4}
In this section, we first evaluate the quality of the individual summaries from the C2F-FAR model and ChatGPT. Then we show the results based on the proposed hybrid summarization pipeline. To measure the performance of C2F-FAR, we mainly rely on automated evaluation metrics such as ROUGE \cite{Lin2004ROUGE}, BERTScore \cite{Zhang2019BERTScore}, BLANC \cite{Vasilyev2020Fill} and ESTIME \cite{Vasilyev2021ESTIME,Vasilyev2021Consistency}. To measure the quality of ChatGPT output, we also turn to human evaluations aimed at identifying potential problems related to text coherence, faithfulness, and style. We analyze the results by addressing the question of whether the proposed pipeline could serve as a basis for human editors to perform summarization tasks more efficiently.

\subsection{C2F-FAR Extractive Summarization}
\label{sec:4.1}
We applied the default C2F-FAR setting to the cleaned 20 articles. The articles were distinguished according to the default categories, i.e., 10 documents in business articles and 10 documents in economic reports. We calculated ROUGE, BERTScore, BLANC, and ESTIME scores to get a first insight into how the extractive summaries generated by the model perform compared to the human-written summaries. Table \ref{Tab1} shows the measured scores as well as the length of the summaries in terms of the number of words. It can be seen that the ROUGE-1 results for both document categories are almost comparable to the report of the state-of-the-art models \cite{Pang2022Long} applied to the PubMed and arXiv datasets \cite{Cohan2018A}. However, the ROUGE-2 and ROUGE-L values are much lower than those reported in the literature, which are above 0.2 and 0.4, respectively. This suggests that the summaries produced by C2F-FAR for our dataset can capture the keywords but not the phrases used in the human summaries, although a high BERTScore was obtained. As with the reference-free metrics (Table \ref{Tab2}), the BLANC scores (independent of $\textrm{BLANC}_\textit{help}$ and $\textrm{BLANC}_\textit{tune}$) for both document categories are higher for the model summaries than for the human summaries. This seems reasonable since the generated summaries can be up to 25 sentences long and are thus longer than those written by humans. Given the extractive nature of the model summaries, it also seems reasonable to assume that these summaries can help a language model better understand the original document based on the BLANC definition. Moreover, from the definition of $\textrm{ESTIME}_\textit{alarms}$ and $\textrm{ESTIME}_\textit{soft}$, we can see that the model summaries use far fewer new words than the human summaries and are better adapted to the original text. This is also to be expected since the sentences were extracted directly from the documents using C2F-FAR. Unfortunately, the calculation of the reference-free evaluation metrics used can require a lot of computing time, which limits their application to very long documents such as books.

\begin{table}[H]
    \centering
    \caption{Comparison of C2F-FAR generated and human-written summaries with reference-based metrics. ``Avg'' stands for the average and ``Std'' stands for the standard deviation. ``B'' stands for the business articles and ``E'' stands for the economic reports.}
    \label{Tab1}      
    \begin{tabular}{ccccccc}
	\toprule
	   & ROUGE-1 & ROUGE-2 & ROUGE-L & BERTScore & Words & Words \\
          & & & & & (Human) & (Model) \\
	\midrule
	Avg (B) & 0.45 & 0.10 & 0.17 & 0.84 & 363 & 560 \\
        Std (B) & 0.04 & 0.03 & 0.02 & 0.01 & 28 & 67 \\
	\midrule  
	Avg (E) & 0.44 & 0.11 & 0.16 & 0.84 & 605 & 638 \\
        Std (E) & 0.06 & 0.02 & 0.02 & 0.01 & 503 & 178 \\
	\bottomrule 
    \end{tabular}
\end{table} 

\begin{table}[H]
    \centering
    \caption{Comparison of C2F-FAR generated and human-written summaries with reference-free metrics. The definitions of ``Avg'', ``Std'', ``B'', and ``E'' are the same as in Table \ref{Tab1}.}
    \label{Tab2}      
    \begin{tabular}{ccccccccccc}
	\toprule
	 & BLANC & BLANC & BLANC & BLANC & ESTIME & ESTIME & ESTIME & ESTIME \\
      & \textit{help} & \textit{help} & \textit{tune} & \textit{tune} & \textit{alarms} & \textit{alarms} & \textit{soft} & \textit{soft} \\
      & (Human) & (Model) & (Human) & (Model) & (Human) & (Model) & (Human) & (Model) \\
	\midrule
	Avg (B) & 0.11 & 0.19 & 0.06 & 0.15 & 187 & 26 & 0.61 & 0.97 \\
        Std (B) & 0.04 & 0.05 & 0.04 & 0.05 & 26 & 6 & 0.05 & 0.01 \\
	\midrule  
	Avg (E) & 0.12 & 0.17 & 0.05 & 0.14 & 292 & 24 & 0.65 & 0.98 \\
        Std (E) & 0.02 & 0.04 & 0.02 & 0.05 & 262 & 9 & 0.04 & 0.01 \\
	\bottomrule 
    \end{tabular}
\end{table} 

We also performed a human evaluation on a sample pair of model and human summaries based on Article 25305. In the appendix \ref{app:B}, we used different colored text to highlight the semantic overlap between the two summaries. It can be seen that the original C2F-FAR method can already capture some of the most important facts that appear in the human summary (and thus in the original document). This has led us to believe that C2F-FAR can be used as an extractive basis for the hybrid summarization pipeline. Nevertheless, challenges remain in applying C2F-FAR to the summarization of long documents in practice: (1) The computation time is relatively long. (2) Summaries consist of parts of sentences that may not contain all the important information and structure of the original document, which prevents a good overview of the original document. (3) There are some model parameters that should be adjusted and may affect the summarization results. 

\subsection{ChatGPT Abstractive Summarization}
\label{sec:4.2}
\begin{figure}[H]
    \centering
    \includegraphics[width=6.5in]{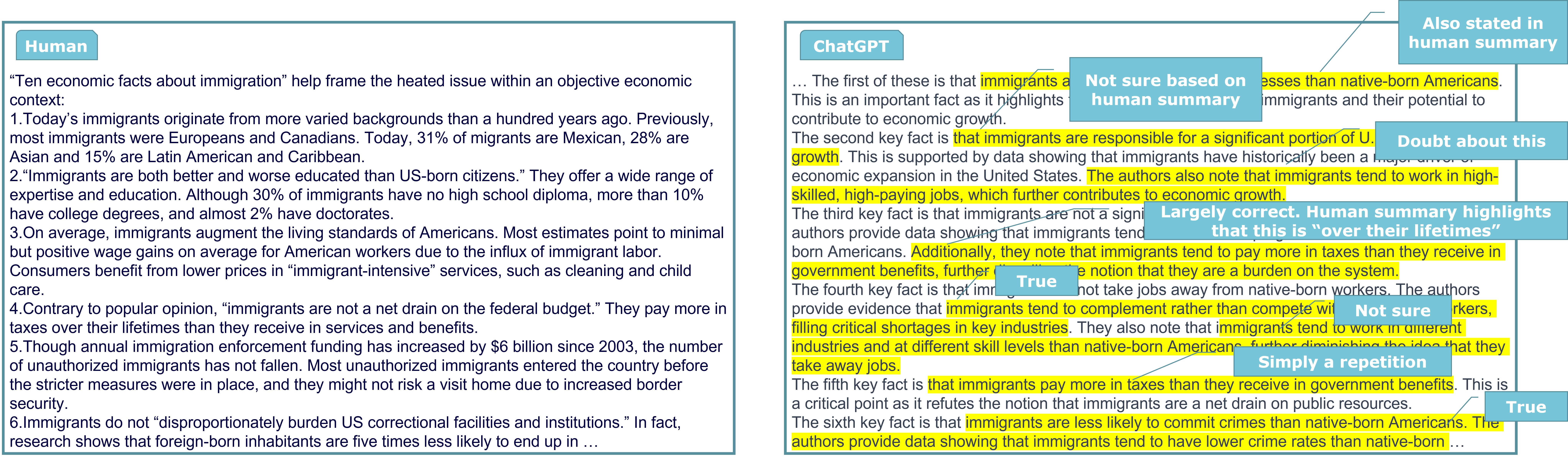}
    \caption{Comparison of the ChatGPT summary with the human summary of the ``Ten Economic Facts about Immigration'' article. The yellow text highlights the possible matches and discrepancies in the ChatGPT summary.}
    \label{Fig3}
\end{figure}

To understand ChatGPT's ability to summarize long documents, we applied the prompt used in Figure \ref{Fig2} to five business articles in the dataset. Figure \ref{Fig3} compares the model and human summaries using the same article ``Ten Economic Facts about Immigration'' and highlights the potential problems of model summarization using a human evaluation. We also had ChatGPT directly summarize some English books in the dataset and examine the quality of the summaries using human evaluation. Unfortunately, the detailed human annotations of the model summaries compared to the human-written summaries cannot be published due to book rights restrictions. 

Based on these experiments, we came to the following conclusions: (1) ChatGPT can generate long summaries very quickly, which favors a practical application of this technology. (2) The generated summaries follow a very mechanical and sometimes repetitive structure (e.g., ``The book begins with ...'') that is independent of the given instruction. (3) The wording style of the summaries can be controlled to some extent, taking into account the given prompt. (4) The content is sometimes somewhat general as there are no specific words or numbers from the book, while the human-written summary contains a very specific description of the book with its typical vocabulary, numbers and specific examples. (5) Very often, the facts in the summaries generated by ChatGPT do not match the original documents and the human summaries, which can only be detected with certainty by human evaluations with detailed knowledge of the original document. (6) Sometimes a book with the same title but different authors is found and summarized, which is wrong. These observations promoted us to further input the sentences extracted by C2F-FAR into ChatGPT in the hope of improving the faithfulness of the summaries generated by the model.

\subsection{C2F-FAR and ChatGPT Hybrid Summarization}
\label{sec:4.3}
In the proposed hybrid summarization pipeline, C2F-FAR was used to extract the important sentences from the original document. The order of the extracted sentences follows the order in which they appear in the document. The input of C2F-FAR results into ChatGPT aims to improve the consistency of the facts in the generated summaries with the original document. ChatGPT was also used for paraphrasing to improve text coherence and style. 

\begin{table}[H]
    \centering
    \caption{Testing the paraphrasing ability of ChatGPT against the C2F-FAR extracted sentences using the 20 cleaned articles. The definitions of ``Avg'' and ``Std'' are the same as in Table \ref{Tab1}.}
    \label{Tab3}      
    \begin{tabular}{ccccccccc}
	\toprule
	   & ROUGE-1 & ROUGE-2 & ROUGE-L & BERTScore & BLANC & BLANC & ESTIME & ESTIME \\
          & & & & & \textit{help} & \textit{tune} & \textit{alarms} & \textit{soft} \\
	\midrule
        \textbf{Human} & & & & & & & & \\
        \midrule
	Avg & & & & & 0.11 & 0.05 & 238 & 0.64 \\
        Std & & & & & 0.03 & 0.03 & 197 & 0.05 \\
	\midrule  
        \textbf{C2F-FAR} & & & & & & & & \\
        \midrule  
	Avg & 0.44 & 0.11 & 0.17 & 0.84 & 0.17 & 0.13 & 27 & 0.97 \\
        Std & 0.05 & 0.03 & 0.02 & 0.01 & 0.05 & 0.05 & 8.72 & 0.01 \\
        \midrule  
        \textbf{ChatGPT} & & & & & & & & \\
        \midrule 
        Avg & 0.42 & 0.11 & 0.17 & 0.85 & 0.15 & 0.10 & 199 & 0.81 \\
        Std & 0.05 & 0.03 & 0.02 & 0.01 & 0.04 & 0.04 & 98 & 0.05 \\
	\bottomrule 
    \end{tabular}
\end{table} 

First, we tested ChatGPT's paraphrasing ability on the input sentences using the following examples. Based on the 20 cleaned articles, we had C2F-FAR extract the first 25 most important sentences from each document and had ChatGPT paraphrase this text. The prompt was ``Please rewrite the text in your own words''. The 25 sentences selected were reviewed to cover the full range of the document. Based on the 20 cleaned books, we had C2F-FAR extract the first 100 most important sentences from each document and had ChatGPT either summarize, paraphrase or summarize and paraphrase this text in about 25 sentences. The prompt was ``Please summarize (rewrite, summarize and rewrite) the text in your own words in about 25 sentences''. Note that 100 sentences are far from covering the entire content of the original book and are only used as test cases. Again, summarization results were assessed using both automated metrics and human evaluations. 

\begin{table}[H]
    \centering
    \caption{Testing the paraphrasing ability of ChatGPT using the 20 cleaned books. The definitions of ``Avg'' and ``Std'' are the same as in Table \ref{Tab1}. ``S100->25'' means that the 100 sentences extracted by C2F-FAR are summarized into 25 sentences. ``R100->25'' means that the results of ``S100->25'' are rewritten in ChatGPT's own words. ``SR100->25'' summarizes the 100 sentences extracted by C2F-FAR and rewrites them into 25 sentences.}
    \label{Tab4}      
    \begin{tabular}{ccccccccc}
	\toprule
	   & ROUGE-1 & ROUGE-2 & ROUGE-L & BERTScore & BLANC & BLANC & ESTIME & ESTIME \\
          & & & & & \textit{help} & \textit{tune} & \textit{alarms} & \textit{soft} \\
	\midrule
        \textbf{S100->25} & & & & & & & & \\
        \midrule  
	Avg & 0.48 & 0.19 & 0.23 & 0.84 & 0.16 & 0.10 & 165 & 0.55 \\
        Std & 0.08 & 0.09 & 0.07 & 0.01 & 0.06 & 0.05 & 38 & 0.11 \\
        \midrule  
        \textbf{R100->25} & & & & & & & & \\
        \midrule 
        Avg & 0.44 & 0.13 & 0.19 & 0.83 & 0.13 & 0.08 & 167 & 0.51 \\
        Std & 0.06 & 0.06 & 0.05 & 0.01 & 0.06 & 0.04 & 42 & 0.11 \\
        \midrule  
        \textbf{SR100->25} & & & & & & & & \\
        \midrule 
        Avg & 0.48 & 0.19 & 0.24 & 0.84 & 0.16 & 0.11 & 147 & 0.56 \\
        Std & 0.07 & 0.09 & 0.09 & 0.01 & 0.07 & 0.04 & 34 & 0.11 \\
	\bottomrule 
    \end{tabular}
\end{table} 

Table \ref{Tab3} shows both reference-based and reference-free metrics used to score the summaries for articles. Summaries written by humans were considered the golden standard. On the one hand, ChatGPT paraphrases the sentences extracted by C2F-FAR, resulting in a slightly lower score for ROUGE-1 and BLANC and a slightly higher BERTScore compared to the summaries generated by C2F-FAR. ROUGE-2 and ROUGE-L remain the same after paraphrasing, suggesting that ChatGPT relies heavily on the input text (despite its paraphrasing ability) to approximate  the human summaries. On the other hand, paraphrasing introduces many new words to represent the same semantic meaning of the original document. This results in a higher $\textrm{ESTIME}_\textit{alarms}$ and a lower $\textrm{ESTIME}_\textit{soft}$ value compared to the C2F-FAR generated summaries. Nevertheless, the $\textrm{ESTIME}_\textit{alarms}$ value for the ChatGPT output is still lower than for the human-written summaries, and the $\textrm{ESTIME}_\textit{soft}$ value for the ChatGPT output is still higher than for the human-written summaries. Overall, we have observed that ChatGPT can rewrite the input text to some extent. 

Table \ref{Tab4} also shows both reference-based and reference-free metrics that were used to evaluate the summaries for books. Summaries formed from the top 25 sentences using C2F-FAR were considered the ``gold standard''. This ``gold standard'' is more of a reference, as we are aware that the information contained in the top 25 sentences may not be completely the same as that contained in the top 100 sentences. It can be seen that the difference between the prompts ``Summarize the text in your own words'' (S100->25) and ``Summarize and rewrite the text in your own words'' (SR100->25) has almost no effect on the summarization results, except that SR100->25 corresponds to a slightly lower value of $\textrm{ESTIME}_\textit{alarms}$. In contrast, the two-step prompts ``Summarize the text in your own words'' (S100->25) and ``Rewrite the text in your own words'' (R100->25) result in more changes to the text, which is particularly evident in the decrease in ROUGE and BLANC scores. These tests give an insight into the paraphrasing capability of ChatGPT as a function of the prompt and its potential usefulness in the hybrid summarization pipeline.

Second, to complete the evaluation of the summaries produced by our hybrid summarization pipeline, we conducted thorough human evaluations and even compared the summaries produced by GPT-3.5 and GPT-4 using books. The following observations can be made. (1) Coherence: Because the input to ChatGPT is long, smaller chunks are required to divide the long document. However, due to the very mechanical structure of the text generated by ChatGPT, the coherence between the texts generated by different chunks is not ideal. Sometimes there is even a lack of coherence within the paragraph. (2) Faithfulness: The summaries produced by ChatGPT appear to be correct at the beginning. Halfway through the text, the machine begins to hallucinate more and more, which is accompanied by other problems such as illogical conclusions, redundancies, literal repetitions and inexplicable abbreviations. These problems are significantly improved with the GPT-4 model, which summarizes the same document with the same pipeline. (3) Style: The summaries created with ChatGPT have a very mechanical structure that is more like a running account of facts based on the C2F-FAR extracted texts. In addition, they do not contain subheadings, quotations, metaphors, idiomatic expressions or puns that make the text unique and interesting. Only the text created with GPT-4 has some of these stylistic features. 

Overall, the summaries produced by ChatGPT look impressive at first glance. A detailed evaluation shows that they need to be thoroughly checked for text coherence, faithfulness and style before they can be used as inspiration material for human editors. Among these problems, we believe that text faithfulness and style are the most critical issues that may hinder the efficient application of the proposed summarization pipeline in production. So far, neither intrinsic nor extrinsic hallucinations can be easily and comprehensively found using automated evaluation metrics based on information extraction, natural language inference and question answering \cite{Ji2023Survey}. Since the tolerance for hallucination in professional text summarization is very low, a great deal of effort is still required on the human side. Moreover, human editors have to paraphrase the machine-generated texts in their own words to match the style of the original document and the specific requirements of the readers. So far, there is a lack of guidelines to help editors use the machine-generated materials more efficiently. Therefore, ChatGPT and GPT-4 can currently only be used as a support tool for appropriate tasks. It would be risky to rely on them completely in a professional setting. Nevertheless, they are promising when it comes to creatively expanding human capabilities.

\section{Conclusions and Outlook}
\label{sec:5}
In this work, we conducted a practical study of hybrid summarization of long documents using C2F-FAR and ChatGPT to better understand the importance of such a summarization pipeline for professional use. We used C2F-FAR's ability to extract important sentences from texts and ChatGPT's zero-shot ability to summarize and paraphrase texts. The results measured against the existing automated evaluation metrics suggest that the machine-generated summaries look as good as those written by humans, especially when the ROUGE-1 score is taken into account. However, several critical issues were identified in the human evaluation in terms of text coherence, faithfulness and style. Most importantly, such powerful models as ChatGPT and GPT-4 can still produce unfaithful and very mechanical summaries given the input texts that need to be carefully checked and rewritten by human editors before they are used in practice. This shows that existing LLMs cannot yet fully replace the role of human effort in composing accurate and correct texts. However, based on our experience in this project, we believe that ChatGPT, especially GPT-4, provides faster, better and more inspiring results in text summarization compared to the other state-of-the-art models. A future area of work should focus on how to effectively and efficiently assess the quality of long document summaries in the production environment and how to use the model-generated texts to improve human work efficiency.

\section*{Acknowledgments}
\label{Ack}
The authors thank Innosuisse, Switzerland (Project Number 63434.1 INNO-SBM) for financial support of this work.

\bibliographystyle{unsrtnat}
%\bibliography{references}  %%% Uncomment this line and comment out the ``thebibliography'' section below to use the external .bib file (using bibtex) .


\begin{thebibliography}{1}

\bibitem{ElKassas2021Automatic} El-Kassas, W. S., Salama, C. R., Rafea, A. A., \& Mohamed, H. K. (2021). Automatic text summarization: A comprehensive survey. Expert Systems with Applications, 165, 113679.
\bibitem{Pilault2020On} Pilault, J., Li, R., Subramanian, S., \& Pal, C. (2020, November). On extractive and abstractive neural document summarization with transformer language models. In Proceedings of the 2020 Conference on Empirical Methods in Natural Language Processing (EMNLP) (pp. 9308-9319).
\bibitem{Cajueiro2023A} Cajueiro, D. O., Nery, A. G., Tavares, I., De Melo, M. K., Reis, S. A. D., Weigang, L., \& Celestino, V. R. (2023). A comprehensive review of automatic text summarization techniques: method, data, evaluation and coding. arXiv preprint arXiv:2301.03403.
\bibitem{Zhao2023A} Zhao, W. X., Zhou, K., Li, J., Tang, T., Wang, X., Hou, Y., ... \& Wen, J. R. (2023). A Survey of large language models. arXiv preprint arXiv:2303.18223.
\bibitem{Yang2023Harnessing} Yang, J., Jin, H., Tang, R., Han, X., Feng, Q., Jiang, H., ... \& Hu, X. (2023). Harnessing the power of LLMs in practice: A survey on ChatGPT and beyond. arXiv preprint arXiv:2304.13712.
\bibitem{Zhang2023Benchmarking} Zhang, T., Ladhak, F., Durmus, E., Liang, P., McKeown, K., \& Hashimoto, T. B. (2023). Benchmarking large language models for news summarization. arXiv preprint arXiv:2301.13848.
\bibitem{Fabbri2021SummEval} Fabbri, A. R., Kryściński, W., McCann, B., Xiong, C., Socher, R., \& Radev, D. (2021). SummEval: Re-evaluating summarization evaluation. Transactions of the Association for Computational Linguistics, 9, 391-409.
\bibitem{Koh2022An} Koh, H. Y., Ju, J., Liu, M., \& Pan, S. (2022). An empirical survey on long document summarization: datasets, models, and metrics. ACM Computing Surveys, 55(8), 1-35.
\bibitem{Pang2022Long} Pang, B., Nijkamp, E., Kryściński, W., Savarese, S., Zhou, Y., \& Xiong, C. (2022). Long document summarization with top-down and bottom-up inference. arXiv preprint arXiv:2203.07586.
\bibitem{Stiennon2020Learning} Stiennon, N., Ouyang, L., Wu, J., Ziegler, D., Lowe, R., Voss, C., ... \& Christiano, P. F. (2020). Learning to summarize with human feedback. Advances in Neural Information Processing Systems, 33, 3008-3021.
\bibitem{Wu2021Recursively} Wu, J., Ouyang, L., Ziegler, D. M., Stiennon, N., Lowe, R., Leike, J., \& Christiano, P. (2021). Recursively summarizing books with human feedback. arXiv preprint arXiv:2109.10862.
\bibitem{Ouyang2022Training} Ouyang, L., Wu, J., Jiang, X., Almeida, D., Wainwright, C., Mishkin, P., ... \& Lowe, R. (2022). Training language models to follow instructions with human feedback. Advances in Neural Information Processing Systems, 35, 27730-27744.
\bibitem{Kryściński2019Neural} Kryściński, W., Keskar, N. S., McCann, B., Xiong, C., \& Socher, R. (2019). Neural text summarization: A critical evaluation. arXiv preprint arXiv:1908.08960.
\bibitem{Zhang2023A} Zhang, C., Zhang, C., Zheng, S., Qiao, Y., Li, C., Zhang, M., ... \& Hong, C. S. (2023). A Complete survey on generative AI (AIGC): Is ChatGPT from GPT-4 to GPT-5 all you need?. arXiv preprint arXiv:2303.11717.
\bibitem{Cao2023A} Cao, Y., Li, S., Liu, Y., Yan, Z., Dai, Y., Yu, P. S., \& Sun, L. (2023). A comprehensive survey of AI-generated content (AIGC): A history of generative AI from GAN to ChatGPT. arXiv preprint arXiv:2303.04226.
\bibitem{Bang2023A} Bang, Y., Cahyawijaya, S., Lee, N., Dai, W., Su, D., Wilie, B., ... \& Fung, P. (2023). A multitask, multilingual, multimodal evaluation of ChatGPT on reasoning, hallucination, and interactivity. arXiv preprint arXiv:2302.04023.
\bibitem{Zhang2023One} Zhang, C., Zhang, C., Li, C., Qiao, Y., Zheng, S., Dam, S. K., ... \& Hong, C. S. (2023). One small step for generative AI, one giant leap for AGI: A complete survey on ChatGPT in AIGC era. arXiv preprint arXiv:2304.06488.
\bibitem{Soni2023Comparing} Soni, M., \& Wade, V. (2023). Comparing abstractive summaries generated by ChatGPT to real summaries through blinded reviewers and text classification algorithms. arXiv preprint arXiv:2303.17650.
\bibitem{Yang2023Exploring} Yang, X., Li, Y., Zhang, X., Chen, H., \& Cheng, W. (2023). Exploring the limits of ChatGPT for query or aspect-based text summarization. arXiv preprint arXiv:2302.08081.
\bibitem{Lin2004ROUGE} Lin, C. Y. (2004, July). ROUGE: A package for automatic evaluation of summaries. In Text Summarization Branches Out (pp. 74-81).
\bibitem{Zhang2023Extractive} Zhang, H., Liu, X., \& Zhang, J. (2023). Extractive summarization via ChatGPT for faithful summary generation. arXiv preprint arXiv:2304.04193.
\bibitem{Wang2023Cross} Wang, J., Liang, Y., Meng, F., Li, Z., Qu, J., \& Zhou, J. (2023). Cross-lingual summarization via ChatGPT. arXiv preprint arXiv:2302.14229.
\bibitem{Ma2023ImpressionGPT} Ma, C., Wu, Z., Wang, J., Xu, S., Wei, Y., Liu, Z., ... \& Li, X. (2023). ImpressionGPT: An iterative optimizing framework for radiology report summarization with ChatGPT. arXiv preprint arXiv:2304.08448.
\bibitem{Luo2023Chatgpt} Luo, Z., Xie, Q., \& Ananiadou, S. (2023). ChatGPT as a factual inconsistency evaluator for abstractive text summarization. arXiv preprint arXiv:2303.15621.
\bibitem{Liu2023GPTEval} Liu, Y., Iter, D., Xu, Y., Wang, S., Xu, R., \& Zhu, C. (2023). GPTEval: NLG evaluation using GPT-4 with better human alignment. arXiv preprint arXiv:2303.16634.
\bibitem{Gao2023Human} Gao, M., Ruan, J., Sun, R., Yin, X., Yang, S., \& Wan, X. (2023). Human-like summarization evaluation with ChatGPT. arXiv preprint arXiv:2304.02554.
\bibitem{Wang2023Is} Wang, J., Liang, Y., Meng, F., Shi, H., Li, Z., Xu, J., ... \& Zhou, J. (2023). Is ChatGPT a good NLG evaluator? A preliminary study. arXiv preprint arXiv:2303.04048.
\bibitem{Liang2022An} Liang, X., Li, J., Wu, S., Zeng, J., Jiang, Y., Li, M., \& Li, Z. (2022). An efficient coarse-to-fine facet-aware unsupervised summarization framework based on semantic blocks. arXiv preprint arXiv:2208.08253.
\bibitem{Bubeck2023Sparks} Bubeck, S., Chandrasekaran, V., Eldan, R., Gehrke, J., Horvitz, E., Kamar, E., ... \& Zhang, Y. (2023). Sparks of artificial general intelligence: Early experiments with GPT-4. arXiv preprint arXiv:2303.12712.
\bibitem{Liu2023Summary} Liu, Y., Han, T., Ma, S., Zhang, J., Yang, Y., Tian, J., ... \& Ge, B. (2023). Summary of ChatGPT/GPT-4 research and perspective towards the future of large language models. arXiv preprint arXiv:2304.01852.
\bibitem{Zhang2022HEGEL} Zhang, H., Liu, X., \& Zhang, J. (2022). HEGEL: Hypergraph transformer for long document summarization. arXiv preprint arXiv:2210.04126.
\bibitem{Liu2021HETFORMER} Liu, Y., Zhang, J. G., Wan, Y., Xia, C., He, L., \& Yu, P. S. (2021). HETFORMER: Heterogeneous transformer with sparse attention for long-text extractive summarization. arXiv preprint arXiv:2110.06388.
\bibitem{Ruan2022HiStruct} Ruan, Q., Ostendorff, M., \& Rehm, G. (2022). HiStruct+: Improving extractive text summarization with hierarchical structure information. arXiv preprint arXiv:2203.09629.
\bibitem{Cho2022Toward} Cho, S., Song, K., Wang, X., Liu, F., \& Yu, D. (2022). Toward unifying text segmentation and long document summarization. arXiv preprint arXiv:2210.16422.
\bibitem{Bian2022GoSum} Bian, J., Huang, X., Zhou, H., \& Zhu, S. (2022). GoSum: Extractive summarization of long documents by reinforcement learning and graph organized discourse state. arXiv preprint arXiv:2211.10247.
\bibitem{Xie2022GRETEL} Xie, Q., Huang, J., Saha, T., \& Ananiadou, S. (2022). GRETEL: Graph contrastive topic enhanced language model for long document extractive summarization. arXiv preprint arXiv:2208.09982.
\bibitem{Gu2021MemSum} Gu, N., Ash, E., \& Hahnloser, R. H. (2021). MemSum: Extractive summarization of long documents using multi-step episodic Markov decision processes. arXiv preprint arXiv:2107.08929.
\bibitem{Cohan2018A} Cohan, A., Dernoncourt, F., Kim, D. S., Bui, T., Kim, S., Chang, W., \& Goharian, N. (2018). A discourse-aware attention model for abstractive summarization of long documents. arXiv preprint arXiv:1804.05685.
\bibitem{Cao2022HIBRIDS} Cao, S., \& Wang, L. (2022). HIBRIDS: Attention with hierarchical biases for structure-aware long document summarization. arXiv preprint arXiv:2203.10741.
\bibitem{Zhang2021SummN} Zhang, Y., Ni, A., Mao, Z., Wu, C. H., Zhu, C., Deb, B., ... \& Zhang, R. (2021). Summ$^{N}$: A multi-stage summarization framework for long input dialogues and documents. arXiv preprint arXiv:2110.10150.
\bibitem{Mao2021DYLE} Mao, Z., Wu, C. H., Ni, A., Zhang, Y., Zhang, R., Yu, T., ... \& Radev, D. (2021). DYLE: Dynamic latent extraction for abstractive long-input summarization. arXiv preprint arXiv:2110.08168.
\bibitem{Phang2022Investigating} Phang, J., Zhao, Y., \& Liu, P. J. (2022). Investigating efficiently extending transformers for long input summarization. arXiv preprint arXiv:2208.04347.
\bibitem{He2022zCode} He, P., Peng, B., Lu, L., Wang, S., Mei, J., Liu, Y., ... \& Huang, X. (2022). Z-Code++: A pre-trained language model optimized for abstractive summarization. arXiv preprint arXiv:2208.09770.
\bibitem{Huang2021Efficient} Huang, L., Cao, S., Parulian, N., Ji, H., \& Wang, L. (2021). Efficient attentions for long document summarization. arXiv preprint arXiv:2104.02112.
\bibitem{Kryscinski2021BookSum} Kryściński, W., Rajani, N., Agarwal, D., Xiong, C., \& Radev, D. (2021). BookSum: A collection of datasets for long-form narrative summarization. arXiv preprint arXiv:2105.08209.
\bibitem{Ji2023Survey} Ji, Z., Lee, N., Frieske, R., Yu, T., Su, D., Xu, Y., ... \& Fung, P. (2023). Survey of hallucination in natural language generation. ACM Computing Surveys, 55(12), 1-38.
\bibitem{Ladhak2020Exploring} Ladhak, F., Li, B., Al-Onaizan, Y., \& McKeown, K. (2020). Exploring content selection in summarization of novel chapters. arXiv preprint arXiv:2005.01840.
\bibitem{Koh2022How} Koh, H. Y., Ju, J., Zhang, H., Liu, M., \& Pan, S. (2022). How far are we from robust long abstractive summarization?. arXiv preprint arXiv:2210.16732.
\bibitem{Wang2022SQuALITY} Wang, A., Pang, R. Y., Chen, A., Phang, J., \& Bowman, S. R. (2022). SQuALITY: Building a long-document summarization dataset the hard way. arXiv preprint arXiv:2205.11465.
\bibitem{Zhang2019BERTScore} Zhang, T., Kishore, V., Wu, F., Weinberger, K. Q., \& Artzi, Y. (2019). BERTScore: Evaluating text generation with BERT. arXiv preprint arXiv:1904.09675.
\bibitem{Zhao2019MoverScore} Zhao, W., Peyrard, M., Liu, F., Gao, Y., Meyer, C. M., \& Eger, S. (2019). MoverScore: Text generation evaluating with contextualized embeddings and earth mover distance. arXiv preprint arXiv:1909.02622.
\bibitem{Vasilyev2020Fill} Vasilyev, O., Dharnidharka, V., \& Bohannon, J. (2020). Fill in the BLANC: Human-free quality estimation of document summaries. arXiv preprint arXiv:2002.09836.
\bibitem{Vasilyev2021ESTIME} Vasilyev, O., \& Bohannon, J. (2021, November). ESTIME: Estimation of summary-to-text inconsistency by mismatched embeddings. In Proceedings of the 2nd Workshop on Evaluation and Comparison of NLP Systems (pp. 94-103).
\bibitem{Vasilyev2021Consistency} Vasilyev, O., \& Bohannon, J. (2021). Consistency and coherence from points of contextual similarity. arXiv preprint arXiv:2112.11638.
\bibitem{Vig2021SummVis} Vig, J., Kryściński, W., Goel, K., \& Rajani, N. F. (2021). SummVis: Interactive visual analysis of models, data, and evaluation for text summarization. arXiv preprint arXiv:2104.07605.
\bibitem{Gehrmann2022Repairing} Gehrmann, S., Clark, E., \& Sellam, T. (2022). Repairing the cracked foundation: A survey of obstacles in evaluation practices for generated text. arXiv preprint arXiv:2202.06935.





\end{thebibliography}

\begin{appendices}

\section{Names of the Selected Articles and Books in Our Dataset}
\label{app:A}
\paragraph{Articles (Due to rights restrictions, not all articles are listed.)}\mbox{}\\
ID: 19197; Name: Ten Economic Facts about Immigration; Author: Michael Greenstone, \& Adam Looney; Publication date: 01.09.2010.\\
ID: 19776; Name: Cognition: Minding Risks; Author: Mario  Weick, Tim Hopthrow, Dominic Abrams, \& Peter Taylor-Gooby; Publication date: 01.01.2013.\\
ID: 20414; Name: The Economic Impact of Cybercrime and Cyber Espionage; Author: James Lewis, \& Stewart Baker; Publication date: 01.07.2013.\\
ID: 21901; Name: Do Big Banks Have Lower Operating Costs? Author: Anna Kovner, James Vickery, \& Lily Zhou; Publication date: 01.03.2014.\\
ID: 22641; Name: Smart Travel; Author: World Economic Forum; Publication date: 01.06.2014.\\
ID: 24037; Name: Rewarding Interactions; Author: Cindy Faust, Mark Sage, \& Paul Sage; Publication date: 01.12.2014.\\
ID: 24640; Name: Always On, Never Done; Author: Jennifer J. Deal; Publication date: 01.08.2013.\\
ID: 25305; Name: Myths, Exaggerations and Uncomfortable Truths; Author: Carolyn Heller Baird; Publication date: 01.08.2015.\\
ID: 25575; Name: Managing the Risk and Impact of Future Epidemics; Author: World Economic Forum; Publication date: 01.06.2015.\\
ID: 25578; Name: Developing Leadership by Building Psychological Capital; Author: Marian N. Ruderman, \& Cathleen Clerkin; Publication date: 01.08.2015.\\
ID: 25756; Name: A Firm-Level Perspective on the Role of Rents in the Rise in Inequality; Author: Peter Orszag; Publication date: 01.10.2015.\\
ID: 28071; Name: Completing Europe's Economic and Monetary Union; Author: Jean-Claude Juncker; Publication date: 01.06.2015.\\
ID: 28116: Name: The Future of Operational Excellence; Author: Gaurav Bhatnagar, \& Heather Gilmartin Adams; Publication date: 01.12.2015.\\
ID: 28152; Name: Gauging the Ability of the FOMC to Respond to Future Recessions; Author: David Reifschneider; Publication date: 01.08.2016.\\
ID: 29166; Name: Job Satisfaction Index 2016; Author: Krifa, The Happiness Research Institute, TNS Gallup; Publication date: 01.06.2016.\\
ID: 35448; Name: From Great Recession to Great Reshuffling; Author: Economic Innovation Group; Publication date: 01.12.2018.

\paragraph{Books (Due to rights restrictions, not all books are listed.)}\mbox{}\\
ID: 34190; Name: The Execution Factor; Author: Kim Perell; Publication date: 01.09.2018.\\
ID: 38353; Name: From We Will to At Will; Author: Justin  Constantine; Publication date: 01.06.2018.

\section{Human Evaluation of the C2F-FAR and Human Summaries based on Article 25305}
\label{app:B}
The texts of the same colour are semantically related to each other. 

\paragraph{Model Summary}\mbox{}
\textcolor{YellowOrange}{Millennials are the first generation to grow up immersed in a digital world.} \textcolor{Green}{While there are some distinctions among the generations, Millennials attitudes are not poles apart from other employees.} Weve also uncovered three uncomfortable truths that apply to employees of all ages. Lastly, weve made five practical recommendations for helping a multigenerational workforce thrive in todays volatile work environment. \textcolor{Green}{Millennials have similar career aspirations to those of other generations.} Surprisingly, its largely Gen X employees, not Millennials, who think everyone on a successful team should be rewarded. \textcolor{Cyan}{Myth 3: Millennials are digital addicts who want to do and share everything online, without regard for personal or professional boundaries.} Millennials are also quite capable of distinguishing between the personal and professional realms and exercising discretion when they use social media. Many organizations encourage their employees to leverage their personal networks, but this approach could increase the potential for misuse or mistakes if employees don’t have the proper guidance. \textcolor{Red}{Myth 4: Millennials, unlike their older colleagues, cant make a decision without first inviting everyone to weigh in. Despite their reputation for crowdsourcing, Millennials are no more likely than many of their older colleagues to solicit advice at work. True, more than half of all Millennials say they make better business decisions when a variety of people provide input.} Baby Boomers, by contrast, feel far less compelled to include others or worry about seeking consensus and are more skeptical about whether the boss knows best (see Figure 7). Baby Boomers accustomed to making decisions on their own may find it difficult to shift to a more collaborative culture, which can cause tension between older and younger employees. Having the aptitude and tools to do this quickly is essential, as the business landscape becomes more interconnected and complex. Another fiction. \textcolor{Orchid}{When Millennials change jobs, they do so for much the same reasons as Gen X and Baby Boomers. But, as Figure 8 shows, there are no overwhelming generational differences. Seventy-five percent of Millennial respondents said they've held their current positions for three years or more, suggesting that they are no more inclined than older colleagues to gallivant from one job to the next.} In the course of our research, we identified three insights that apply universally and should give business leaders everywhere cause for concern. \textcolor{BrickRed}{More than half of the people we surveyed don't fully understand key elements of their organizations strategy, what they’re supposed to do or what their customers want.} While up to 60 percent of Gen X respondents believe they have a good grasp of these fundamentals, many of their colleagues are struggling (see Figure 9). We asked our respondents to rate their organizations leaders on a number of criteria. We analyzed their answers by role, as well as by generation, to find out how job status influences their perceptions (see How we defined each role). Gen X leaders, in particular, overrate how well they inspire confidence and recognize employees accomplishments (see Figure 10).

\paragraph{Human Summary}\mbox{}
\textcolor{YellowOrange}{Millennials those aged 21 to 34 in 2015 are the first generation to grow up fully immersed in technology.} These digital natives are expected to revolutionize the workforce; older people see them as different from previous generations. Much fear and misconception surrounds millennials; one of the most unkind stereotypes is that they are lazy, entitled, selfish and shallow. However, research tells a different story. \textcolor{Cyan}{Consider five millennial myths, debunked: Their career goals are distinct from those of older generations Like gen X and baby boomers, however, millennials desire financial security and seniority and believe in the value of inspirational leadership and performance-linked recognition. They expect endless praise for their accomplishments In truth, millennials value ethics and fairness more. They are digital addicts who do everything online} Although millennials are good at interacting online, they prefer learning new work-related skills face-to-face. \textcolor{Red}{They make decisions by inviting others to pitch in While 50\% of millennials report valuing group input, they are no more likely than any other generational group to crowdsource advice.} \textcolor{Orchid}{They jump jobs if their current work doesn’t fulfill their passions. In reality, millennials typically change jobs for the same reasons as other generations.} \textcolor{Green}{Millennials as digital natives bring vital value to a work environment in the midst of a digital revolution. But in many ways, they are a lot like their older colleagues.} Firms must recognize some worrying truths about all employees, regardless of generation. \textcolor{BrickRed}{Many employees don’t understand their company’s strategy, what they’re supposed to do or what their customers want.} Leaders need to find creative ways to connect with employees. Most employees believe that their companies handle the customer experience ineffectively and embrace new technologies slowly. Companies should assess the risks of not adopting changes and should introduce innovations to enhance the customer experience. As more millennials have embarked on their careers, expectations of a technological revolution in the workplace have increased. Business leaders must recognize that employees of all ages are complex individuals working in an environment that’s becoming more virtual, more diverse and more volatile by the day and initiate suitable changes in the workplace.

\end{appendices}

\end{document}